%% file: main.tex
\documentclass[10pt,twocolumn,letterpaper]{article}

\usepackage{cvpr}              


\usepackage{amsmath}
\usepackage{amssymb}
\usepackage{geometry}
\usepackage{xcolor}
\usepackage{framed}
\usepackage{tikz}          
\usepackage{tcolorbox}     
\usepackage{enumitem}      
\tcbuselibrary{skins}      

\usepackage{algorithm}
\usepackage{algpseudocode}

\usepackage{booktabs}
\usepackage{multirow}
\usepackage{graphicx}
\usepackage{xcolor}
\usepackage{colortbl}   

\definecolor{cvprblue}{rgb}{0.21,0.49,0.74}
\usepackage[pagebackref,breaklinks,colorlinks,allcolors=cvprblue]{hyperref}

\def\confName{CVPR}
\def\confYear{2026}

\title{\vspace{-0.8cm}GiPL: Generative augmented iterative Pseudo-Labeling for Cross-Domain Few-Shot Object Detection}

\author{
    Jiacong Liu, Shu Luo, Yikai Qin, Yaze Zhao, Yongwei Jiang, Yixiong Zou\thanks{Corresponding author.} \\
    Huazhong University of Science and Technology\\
    {\tt\small \{jiongcongliu, luoshu\_hust, yikaiq, zyaz, jiangyongwei, yixiongz\}@hust.edu.cn}
}

\begin{document}
\maketitle
\input{sec/0_abstract}    
\input{sec/1_intro}
\input{sec/2_related}

\input{sec/3_method}

\input{sec/4_exp}
\input{sec/5_con}

{
    \small
    \bibliographystyle{ieeenat_fullname}
    \bibliography{main}
}


\end{document}

%% file: sec/0_abstract.tex
\begin{abstract}
Vision-language foundation models 
have shown promising zero-shot generalization for Cross-Domain Few-Shot Object Detection (CD-FSOD). However, they face two critical challenges in fine-tuning: insufficient support set utilization due to sparse single-instance annotations, and severe overfitting under extremely limited target-domain samples. 
To address these issues, this paper proposes GiPL, an efficient two-branch training framework.
In the first branch, we design an iterative pseudo-label self-training paradigm, which performs zero-shot inference on the support set to generate reliable pseudo-annotations, fuses them with ground-truth labels, and iteratively optimizes the model to fully exploit support set data. In the second branch, we introduce  generative data augmentation pipeline  using large vision-language models,  
which synthesizes domain-aligned, multi-object annotated images to enrich training samples and suppress overfitting. Extensive experiments on three challenging CD-FSOD datasets (RUOD, CARPK, CarDD) under 1/5/10-shot settings demonstrate that GiPL consistently outperforms state-of-the-art methods 
with significant performance gains.
Code is available at  \href{https://github.com/z-yaz/CDiscover}{CDiscover}.
\end{abstract}

%% file: sec/1_intro.tex
\section{Introduction}
\label{sec:intro}

In recent years, vision-language foundation models (e.g., GroundingDINO \cite{liu2024grounding}, LAE-DINO \cite{LAE-DINO}) have made breakthroughs. Via large-scale multimodal pre-training, they perform well in cross-domain zero-shot and few-shot object detection, providing a new technical path for CD-FSOD \cite{fu2024cdvito}, and can detect unknown-domain objects to a certain extent due to pre-trained vision-language alignment. 
However, adapting these models to new domains with extremely limited labeled samples remains a serious challenge, and CD-FSOD has two urgent core problems to solve.
The first is the contradiction between support set annotation limitations and dense object detection. CD-FSOD support sets usually use instance-level annotation—only one instance is annotated per image even with multiple same-category objects. This causes ``correct detection penalization" during fine-tuning: extra correct annotations are judged as false predictions due to single-instance ground truth, restricting performance improvement. The second is overfitting from data scarcity. Pre-trained vision-language models can only fine-tune on a small number of target-domain support set samples in CD-FSOD. This pre-training-fine-tuning sample imbalance easily leads to overfitting, loss of generalization ability, and even post-fine-tuning performance lower than zero-shot detection in some categories, becoming a key bottleneck for CD-FSOD.

To address these two problems and unlock the potential of vision-language models in CD-FSOD, this paper proposes an efficient, dataset-adaptive two-branch training paradigm. Rather than applying a rigid pipeline, we selectively deploy different branches tailored to the specific characteristics and primary challenges of each target dataset.
For datasets suffering primarily from sparse, single-instance annotations (e.g., dense object scenarios), we adopt the first branch: an iterative pseudo-label self-training paradigm. This branch fine-tunes the pre-trained model, generates reliable pseudo-labels via zero-shot inference on the support set, fuses them with original annotations for further training, and iterates to maximize support set utilization.
Conversely, for datasets where the core bottleneck is severe overfitting due to extreme data scarcity and domain shift, we select the second branch: a generative data augmentation pipeline. Using large language models (i.e., Qwen) and referencing support set images, it generates domain-consistent, multi-object annotated images. These images make up for single-instance annotation deficiencies, enrich training samples, suit low-shot settings, and suppress overfitting.

Comprehensive experiments on three CD-FSOD datasets show the method achieves average performance gains of ~7.63\%, 9.97\% and 10.03\% under 1-shot, 5-shot and 10-shot settings, outperforming baselines and mainstream methods. Our team achieved a score of \textbf{192.7929} and ranked \textbf{second} in the 2nd Cross-Domain Few-Shot Object Detection Challenge~\cite{qiu2026ntire}.

The main contributions are as follows:
\begin{itemize}
\item We propose a dataset-adaptive two-branch CD-FSOD training framework that selectively addresses support set single-instance annotation limitations or data scarcity based on specific dataset characteristics, improving overall detection performance.
\item We design a pseudo-label-based self-training paradigm, which maximizes support set value through  ``fine-tuning --- zero-shot inference --- pseudo-label generation --- fusion training" iteration, enhancing multi-object recognition and localization ability.
\item We propose a generative data augmentation pipeline, using multimodal large language models to generate multi-object, domain-aligned annotated images, alleviating data scarcity-induced overfitting.
\item Extensive experiments on three CD-FSOD benchmarks demonstrate our method achieves state-of-the-art results, validating its effectiveness and generalization.
\end{itemize}

%% file: sec/2_related.tex
\section{Related work}
\label{sec:formatting}

\subsection{General Object Detection}

General object detection has advanced from traditional two-stage and one-stage paradigms to transformer-based detectors. Two-stage methods like Faster R-CNN~\cite{ren2015faster} generate candidate proposals followed by region-wise classification and box regression, while one-stage detectors such as the YOLO series~\cite{redmon2016you,redmon2018yolov3} directly predict objects densely with higher efficiency. DETR~\cite{carion2020end} reformulates detection as set prediction, with variants like Deformable DETR~\cite{zhu2020deformable} and DINO~\cite{zhang2022dino} improving convergence and accuracy. Open-set/open-vocabulary detection extends closed-set settings, with OVR-CNN~\cite{zareian2021open}, ViLD~\cite{gu2021open}, GLIP~\cite{li2022grounded}, and GroundingDINO~\cite{liu2024grounding} enhancing generalization to unseen categories, laying a foundation for few-shot and cross-domain detection.

\subsection{Few-Shot Object Detection}

Few-shot object detection (FSOD) extends few-shot learning~\cite{zhao2026interpretable, fu2023styleadv, zou2024attention, zou2024a, DBLP:conf/cvpr/ZouLH0024} to instance-level detection, requiring novel category recognition and precise localization with limited annotations. Harder than few-shot classification, it tackles foreground-background separation, localization, scale variation, and class imbalance under data scarcity. Early methods build on Faster R-CNN~\cite{ren2015faster}, including Feature Reweighting~\cite{kang2019few}, Meta R-CNN~\cite{yan2019meta}, and TFA~\cite{wang2020frustratingly}. Subsequent works like FSCE~\cite{sun2021fsce} and support-query interaction approaches~\cite{fan2020few} improve adaptation, while recent advances leverage large-scale pretraining and vision-language models for better generalization to novel classes.

\begin{figure*}[t]
    \centering
    \includegraphics[width=\linewidth]{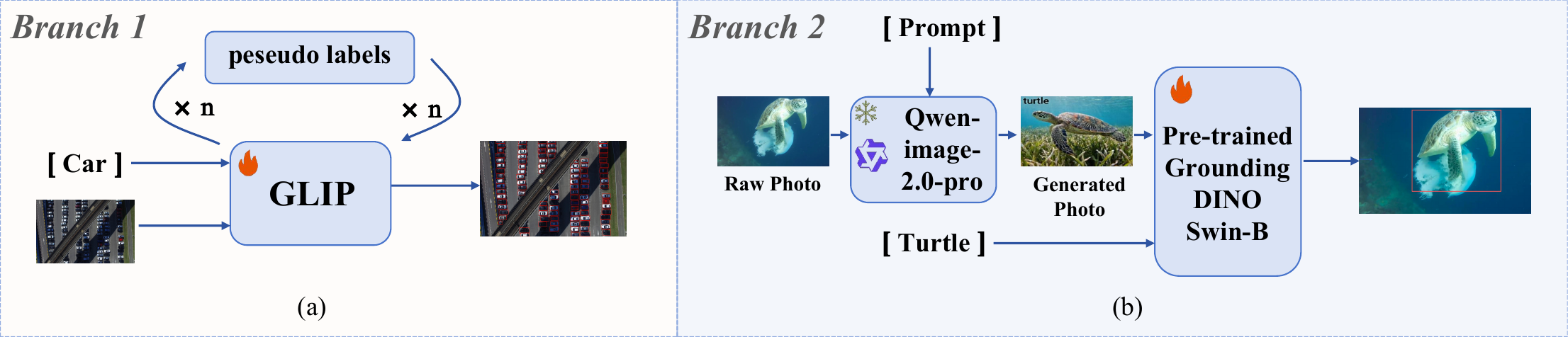}
    \caption{\textbf{Overview of the proposed GiPL-Grounding framework.} 
    Our method addresses the CDFSOD challenge through a dual-branch domain-adaptive strategy. 
    \textbf{Branch 1: Iterative Self-training with Pseudo-labeling (Left).} For CARPK, we employ a cyclic ``inference-annotation-training'' pipeline. A pre-trained detector generates initial pseudo-labels on the target support set, which are then used as supervisory signals to fine-tune the model, progressively correcting localization inaccuracies and alleviating unreasonable background penalties. 
    \textbf{Branch 2: LVLM-based Generative Data Augmentation (Right).} We leverage the multimodal generation capabilities of Qwen to synthesize domain-aligned, semantically consistent training samples. 
    By integrating these two branches, our framework effectively enriches the feature distribution and improves adaptation across diverse target domains.}
    \label{fig:framework}
    \vspace{-0.4cm}
\end{figure*} 

\subsection{Cross-Domain Few-Shot Object Detection}

Cross-domain few-shot object detection (CD-FSOD)\cite{li2025domain} adapts source-domain pretrained detectors to target domains with few labels, facing domain shift and data scarcity. Existing methods include cross-domain alignment approaches like FAFR-CNN, PICA, and DAPN~\cite{fu2024cdvito}; transfer strategies such as AcroFOD, SDAFL, and FUM~\cite{gao2022acrofod,gao2023asyfod}; and integrated frameworks like MoF-SOD, Distill-CDFSOD, and CD-ViTO~\cite{lee2022mofsod,xiong2023cd,fu2024cdvito}. Despite progress, these methods focus on transfer and representation adaptation, neglecting how domain shift disrupts attention and foreground-background discrimination, leaving room for target-aware localization enhancement.

%% file: sec/3_method.tex
\section{Method}

\subsection{Problem Setup}

The Cross-Domain Few-Shot Object Detection (CD-FSOD) task adapts a detector from source domain $D_S$ (distribution $P_S$) to target domain $D_T$ (distribution $P_T \neq P_S$), where each novel class in $C_T$ has few labeled instances. We use the N-way K-shot protocol: support set $S_{N \times K} \subset D_T$ provides K labeled examples per novel class, and query set Q is for evaluation. Adopting the open-source setting from the 2nd Cross-Domain Few-Shot Object Detection Challenge \cite{qiu2026ntire}, we directly fine-tune a pretrained detector (e.g., GroundingDINO \cite{liu2024grounding}) on S and evaluate on Q, instead of pretraining on $D_S$.

\subsection{Revisiting GLIP and GroundingDINO}

GLIP~\cite{li2022grounded} and GroundingDINO~\cite{liu2024grounding} are landmark open-set/multimodal detection models, focusing on vision-language fusion to enable cross-domain and few-shot detection. Let visual feature $\mathbf{V} \in \mathbb{R}^{H \times W \times C}$ (H, W: image dimensions; C: visual feature dimension) and text feature $\mathbf{T} \in \mathbb{R}^{L \times D}$ (L: text length; D: text feature dimension); their core is $\mathbf{V}$-$\mathbf{T}$ alignment.
GLIP reformulates detection as phrase grounding, achieving alignment via contrastive training with fusion loss $\mathcal{L}_{\text{align}} = \text{CE}(\text{sim}(\mathbf{V},\mathbf{T}), y)$ ($\text{sim}(\cdot)$: similarity function; y: matching label). Pre-trained on image-text pairs, it enables arbitrary category detection via text queries but suffers from inter-category interference due to its one-stage architecture and random category name concatenation.

GroundingDINO optimizes GLIP by fusing Transformer-based DINO with grounded pre-training. It uses a dual-encoder-single-decoder architecture, with fusion mechanisms in feature extraction, query initialization, and decoding, and strengthens alignment via the $\text{CrossAttn}(\mathbf{V}, \mathbf{T})$ module. Clause-level text representation and attention masking reduce inter-category interference and improve accuracy.

\subsection{Method Overview}

This paper proposes a novel two-branch training framework (GiPL, Fig.~\ref{fig:framework}), which integrates iterative self-training, pseudo-label generation and large-model data augmentation to address core issues in cross-domain few-shot detection.
The first branch adopts iterative self-training: slightly fine-tuning the pre-trained model first, then inferring on target domain support set samples to generate initial pseudo-labels (bounding boxes and category annotations), and further fine-tuning the model with these pseudo-labels. Cyclic iteration of this process corrects pseudo-label accuracy to form positive feedback, guiding the model to learn target domain features, alleviating unreasonable penalties in dense detection, and improving multi-object adaptability and detection accuracy.
The second branch uses data augmentation to mitigate overfitting from scarce labeled support set data: leveraging LVLMs’ multimodal generation capability, it generates numerous domain-aligned, multi-object annotated images based on a small number of labeled samples. These samples ensure target domain semantic consistency, enrich training diversity, and help the model learn comprehensive features to reduce overfitting.
It should be noted that different branch training is adopted for different datasets.

\subsection{Pseudo-Label Iterative Self-Training}
In the CD-FSOD task,  
only a very small number of bounding box annotations are usually available for each category. This limitation often results in some target objects in the images not being annotated in the initial stage, leading to problems such as low data utilization and sparse annotations. To fully tap the potential value of existing data, make up for the lack of supervision signals  
and effectively improve the model's  cross-domain adaptability, we adopt a pseudo-label iterative self-training strategy.

As shown in Algorithm~\ref{alg:pseudo_label}, the iterative self-training process is carried out in an orderly manner according to the following steps, forming a closed-loop iteration of ``fine-tuning — inference — optimization — retraining": First, we take the limited original labeled data in the support set as the supervision signal to initially fine-tune the pre-trained model, 
enabling it to initially adapt to the target domain features. Then, we use the initially fine-tuned model to perform inference on all samples in the support set, generate pseudo-labels for unlabeled targets, and set a high confidence threshold to retain reliable pseudo-labels while filtering out low-confidence and inaccurate annotations.
Next, we perform strict post-processing steps such as Non-Maximum Suppression (NMS) on the generated pseudo-labels to optimize the accuracy of the pseudo-annotations and eliminate redundant or conflicting labels. Subsequently, we fuse the original real annotations with the optimized high-confidence pseudo-labels to construct an enhanced training set with more comprehensive annotations and larger data volume. Finally, we use this enhanced training set to retrain the initially fine-tuned model, further strengthening the model's ability to learn target domain features.

\begin{algorithm}[t] 
\caption{Iterative Pseudo-Labeling for Object Detection}
\label{alg:pseudo_label}
\begin{algorithmic}[1]
\Require Few-shot annotations $\mathcal{D}_{fs}$, target training images $\mathcal{I}$, detector $\mathcal{M}$, number of rounds $T$, score threshold $\tau_s$, NMS threshold $\tau_n$
\Ensure Pseudo-augmented annotations $\mathcal{D}^{(T)}$

\State $\mathcal{D}^{(0)} \gets \mathcal{D}_{fs}$

\For{$t=1$ to $T$}
    \State Fine-tune $\mathcal{M}$ on $\mathcal{D}^{(t-1)}$
    \State Run $\mathcal{M}$ on $\mathcal{I}$ to obtain predictions $\mathcal{P}^{(t)}$
    \State Filter out predictions with scores lower than $\tau_s$
    \State Apply class-wise NMS with threshold $\tau_n$
    \State Convert the remaining boxes into COCO-style pseudo annotations
    \State Merge them with the original few-shot annotations:
    \[
    \mathcal{D}^{(t)} \gets \mathcal{D}_{fs} \cup \mathcal{P}^{(t)}_{\text{pseudo}}
    \]
\EndFor

\State \Return $\mathcal{D}^{(T)}$ 
\end{algorithmic}
\end{algorithm}

The above process is executed repeatedly in iterations: the model after each round of retraining is used to generate more accurate pseudo-labels, and the optimized pseudo-labels provide more sufficient and reliable supervision signals for the next round of training, forming a positive iterative cycle of ``inference — annotation — training". Through this iterative self-training strategy, the model can gradually learn the feature distribution of the target domain, effectively alleviate the unreasonable penalty problem in dense object detection, significantly improve the adaptability to multi-object scenarios, and enhance the robustness and generalization ability of the model in cross-domain few-shot detection tasks.

\subsection{Generative Augmentation with Qwen}

To mitigate the severe data scarcity and overfitting risks in cross-domain few-shot object detection (CD-FSOD), we adopt a generative data augmentation 
pipeline based on large vision-language models, together with standard detection
augmentations~\cite{pan2025enhance} to enrich the limited support set and improve cross-domain 
generalization.

\begin{figure}[t]
    \centering
    \includegraphics[width=\linewidth]{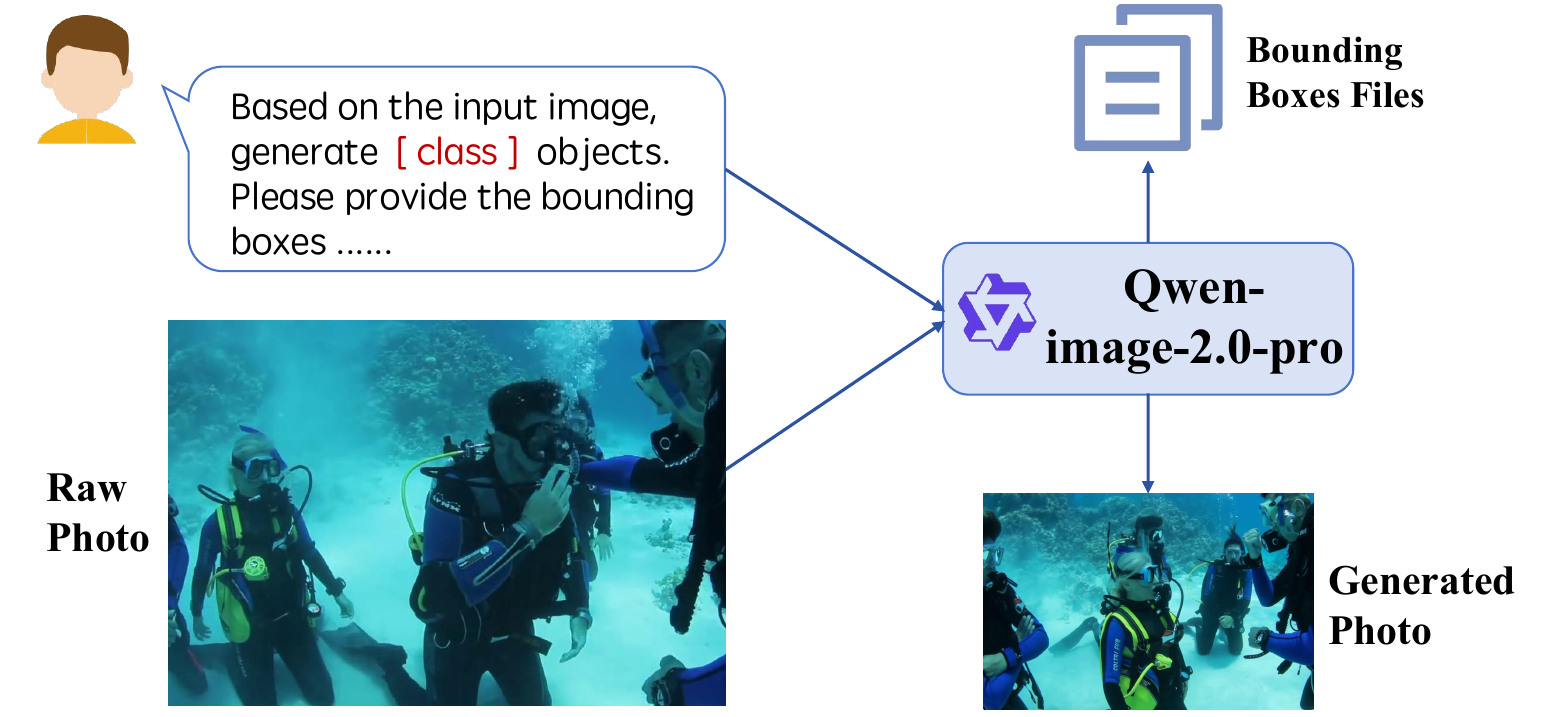}
    \caption{\textbf{Generative data augmentation via Qwen-image-2.0-pro.} 
    As illustrated, given a single support image, the LVLM generates diverse counterparts by simulating various environmental conditions while strictly preserving the object’s category and structural integrity.}
    \label{fig:augmentation}
       \vspace{-0.3cm}
\end{figure}

For Datasets 1 and 3,  as shown in   Fig.~\ref{fig:augmentation},  we employ the Qwen-Image-2.0-pro model to synthesize 
semantically consistent and domain-aligned training samples. Given each support 
image, the model generates new images that preserve the target domain’s style, 
background, and object characteristics while creating valid visual variations. 
This generative augmentation effectively expands 
the samples 
without introducing domain shift, alleviating the label insufficiency in 
1/5/10-shot settings.

\begin{tcolorbox}[
    enhanced,              
     colback=gray!10, 
    colframe=darkgray!80,  
    colbacktitle=lightgray!80, 
    coltitle=black,        
    fonttitle=\normalsize\bfseries, 
    title=Prompt, 
    boxrule=1pt,           
    rounded corners,       
    titlerule=0pt,         
]
Based on the input image, generate [class] objects. Please provide the 
bounding boxes required for object detection based on the positions of the 
generated objects, and save them to a txt file with each object and its 
corresponding box on a single line.  
\end{tcolorbox}

The generation process is guided by textual prompts that enforce category 
consistency and domain fidelity. The prompt used for controlled generation is 
shown in the above. 

%% file: sec/4_exp.tex
\section{Experiments}

\subsection{Experimental Setup}

\textbf{Datasets.} We evaluate our GiPL framework on three diverse target domains that represent significant domain shifts from common detection benchmarks:
\begin{itemize}
    \item \textbf{RUOD:} An underwater object detection dataset characterized by low contrast, color distortion, and varying turbidity.
    \item \textbf{CARPK:} A drone-based parking lot dataset featuring high-density vehicle instances. It often suffers from sparse labeling (only one instance labeled per image) in few-shot settings.
    \item \textbf{CarDD:} A specialized dataset for car damage detection (e.g., dents, scratches), requiring high-precision localization of fine-grained semantic features.
\end{itemize}
\textbf{Evaluation Protocol.} Following the official challenge protocol, we conduct experiments under 1-shot, 5-shot, and 10-shot settings. Performance is measured using Mean Average Precision (mAP) at an IoU threshold of 0.5.
\subsection{Implementation Details}



\noindent \textbf{Model Architectures.} We adopt a domain-adaptive architecture selection strategy to align model inductive biases with target domain attributes. For the dense multi-object scenarios in CARPK~\cite{Carpk}, we utilize 
GLIP-L~\cite{li2022grounded} with a Swin-L backbone. For the specialized domains of RUOD~\cite{fu2023rethinking} and CarDD~\cite{CarDD}, 
we employ Grounding DINO~\cite{liu2024grounding} with a Swin-B backbone, which provides superior localization for fine-grained damage and underwater objects.

\noindent \textbf{Training Configurations.} 
The training protocols are tailored to the specific challenges of each domain. To address the pervasive issue of missing annotations in CARPK's few-shot support set, we integrate an iterative pseudo-labeling workflow, as illustrated in Algorithm~\ref{alg:pseudo_label}, which enables the detector to learn from previously unlabeled vehicle instances. We fine-tune the model for 200 epochs with a learning rate of $5 \times 10^{-5}$ and a batch size of 2, employing an iterative self-training paradigm to refine the model's perception of dense vehicle clusters. In contrast, for the low-resource domains of RUOD and CarDD, we train for 50 epochs with a batch size of 4 on a single NVIDIA A6000 GPU. To mitigate the distribution shift and extreme data scarcity in these domains, we utilize 
Qwen-image-2.0-pro~\cite{wu2025qwen} to synthesize semantically consistent variations of the support set (see Fig.~\ref{fig:augmentation}). This generative augmentation effectively expands the training feature space, providing the model with diverse visual contexts.\\
\noindent \textbf{Inference Configurations.}
During inference, the detector is applied in a standard single-model setting without test-time augmentation. Since the task is object detection, non-maximum suppression (NMS) is used to remove highly overlapped predictions and retain the most confident bounding boxes. We follow the default inference pipeline of the corresponding detector framework for prediction filtering and box selection.

\begin{table}[t]
\centering
\setlength{\abovecaptionskip}{2pt}  
\caption{Performance (mAP@0.5) comparison on three target datasets under 1/5/10-shot settings. GiPL-Grounding (Ours) consistently outperforms the baseline models. The "Avg." column represents the mean performance across all three domains.}
\label{tab:results}
\setlength{\tabcolsep}{3pt} %
\footnotesize
\begin{tabular}{lccccc}
\toprule
\textbf{Method} & \textbf{Shot} & \textbf{RUOD} & \textbf{CARPK} & \textbf{CarDD} & \textbf{Avg.} \\
\midrule
GroundingDINO & 1-shot & 31.5 & 49.0 & 34.2 & 38.2 \\
\textbf{Ours (GiPL)} & 1-shot & \textbf{34.6} & \textbf{63.3} & \textbf{39.7} & \textbf{45.9} \\
\midrule
GLIP & 5-shot & 35.8 & 42.0 & 43.8 & 40.5 \\
\textbf{Ours (GiPL)} & 5-shot & \textbf{41.1} & \textbf{63.0} & \textbf{47.4} & \textbf{50.5} \\
\midrule
GroundingDINO & 10-shot & 34.6 & 40.0 & 47.0 & 40.5 \\
\textbf{Ours (GiPL)} & 10-shot & \textbf{42.1} & \textbf{61.3} & \textbf{48.3} & \textbf{50.6} \\
\bottomrule
\end{tabular}
\vspace{-0.3cm}
\end{table}

\subsection{Main Results}

As shown in Table~\ref{tab:results}, our GiPL-Grounding achieves a final aggregate score of \textbf{192.79}, demonstrating strong generalization across diverse domains.

\textbf{Effectiveness on CARPK.} In the CARPK dataset, our method achieves a remarkable 63.3 mAP in the 1-shot setting, a +14.3 point improvement over the baseline. This surge is attributed to our iterative pseudo-labeling pipeline, which successfully recovers unlabeled vehicle instances, preventing the model from erroneously treating valid targets as background.

\textbf{Robustness in RUOD and CarDD.} For underwater (RUOD) and damage (CarDD) detection, the domain gap is significant. By leveraging Qwen-based generative augmentation, our model learns invariant features despite the low-visibility conditions of RUOD and the subtle visual cues of CarDD. Specifically, for RUOD, our 10-shot performance reaches 42.1 mAP, showing that the synthesis of diverse underwater environments significantly mitigates data scarcity.

\subsection{Ablation study}

Table \ref{tab:ab1} shows that our Qwen-based generative augmentation consistently boosts performance under all few-shot settings, effectively alleviating data scarcity and domain shift.

Table \ref{tab:ab2} illustrates that vanilla fine-tuning suffers from performance degradation caused by sparse annotations, while our pseudo-labeling framework significantly improves detection accuracy on the dense CARPK dataset.

\begin{table}[htbp] 
\centering
\setlength{\abovecaptionskip}{2pt}  
\caption{The 1/5/10-shot ablation study of GroundingDINO on the RUOD and CarDD datasets.} 
\resizebox{0.4\textwidth}{!}{
\begin{tabular}{llccc}
\toprule
& Method & RUOD & CarDD & Avg \\
\midrule 
\multirow{3}{*}{\rotatebox{90}{1-shot}} 
& Zero-shot & 31.5 & 12.0 & 21.75 \\
& + ETS~\cite{pan2025enhance} & 30.8 & 34.2 & 32.50 \\
& \cellcolor{gray!20}+ Augmentation (Ours) 
& \cellcolor{gray!20}\textbf{34.6}  
& \cellcolor{gray!20}\textbf{39.7} 
& \cellcolor{gray!20}\textbf{37.15} \\
\midrule
\multirow{3}{*}{\rotatebox{90}{5-shot}}
& Zero-shot & 31.5 & 12.0 & 21.75 \\
& + ETS~\cite{pan2025enhance} & 35.8 & 43.8 & 39.80 \\
& \cellcolor{gray!20}+ Augmentation (Ours) 
& \cellcolor{gray!20}\textbf{41.1} 
& \cellcolor{gray!20}\textbf{47.4} 
& \cellcolor{gray!20}\textbf{44.25} \\
\midrule
\multirow{3}{*}{\rotatebox{90}{10-shot}}
& Zero-shot & 31.5 & 12.0 & 21.75 \\
& + ETS~\cite{pan2025enhance} & 34.6 & 47.0 & 40.80 \\
& \cellcolor{gray!20}+ Augmentation (Ours) 
& \cellcolor{gray!20}\textbf{42.1} 
& \cellcolor{gray!20}\textbf{48.3} 
& \cellcolor{gray!20}\textbf{45.20} \\
\bottomrule
\end{tabular}
}
\vspace{-0.3cm}
\label{tab:ab1}
\end{table}

\begin{table}[htbp]
\centering
\setlength{\abovecaptionskip}{2pt}  
\caption{The 1/5/10-shot ablation study of GLIP on the CARPK dataset.} 
\resizebox{0.3\textwidth}{!}{
\begin{tabular}{llc}
\toprule
& Method & CARPK \\
\midrule 
\multirow{3}{*}{\rotatebox{90}{1-shot}} 
& Zero-shot  & 50.7 \\
& + fine-tuned  & 49.4 \\
& \cellcolor{gray!20}+ Pseudo-Label (Ours)  
& \cellcolor{gray!20}\textbf{63.3} \\
\midrule
\multirow{3}{*}{\rotatebox{90}{5-shot}}
& Zero-shot  & 50.7 \\
& + fine-tuned  & 42.5 \\
& \cellcolor{gray!20}+ Pseudo-Label (Ours)  
& \cellcolor{gray!20}\textbf{63.0} \\
\midrule
\multirow{3}{*}{\rotatebox{90}{10-shot}}
& Zero-shot  & 50.7 \\
& + fine-tuned  & 40.9 \\
& \cellcolor{gray!20}+ Pseudo-Label (Ours)  
& \cellcolor{gray!20}\textbf{61.3} \\
\bottomrule
\end{tabular}
}
\label{tab:ab2}
\end{table}

%% file: sec/5_con.tex
\subsection{Pseudo-Label Threshold Analysis}

\begin{figure}[t]
    \centering
    \includegraphics[width=\linewidth]{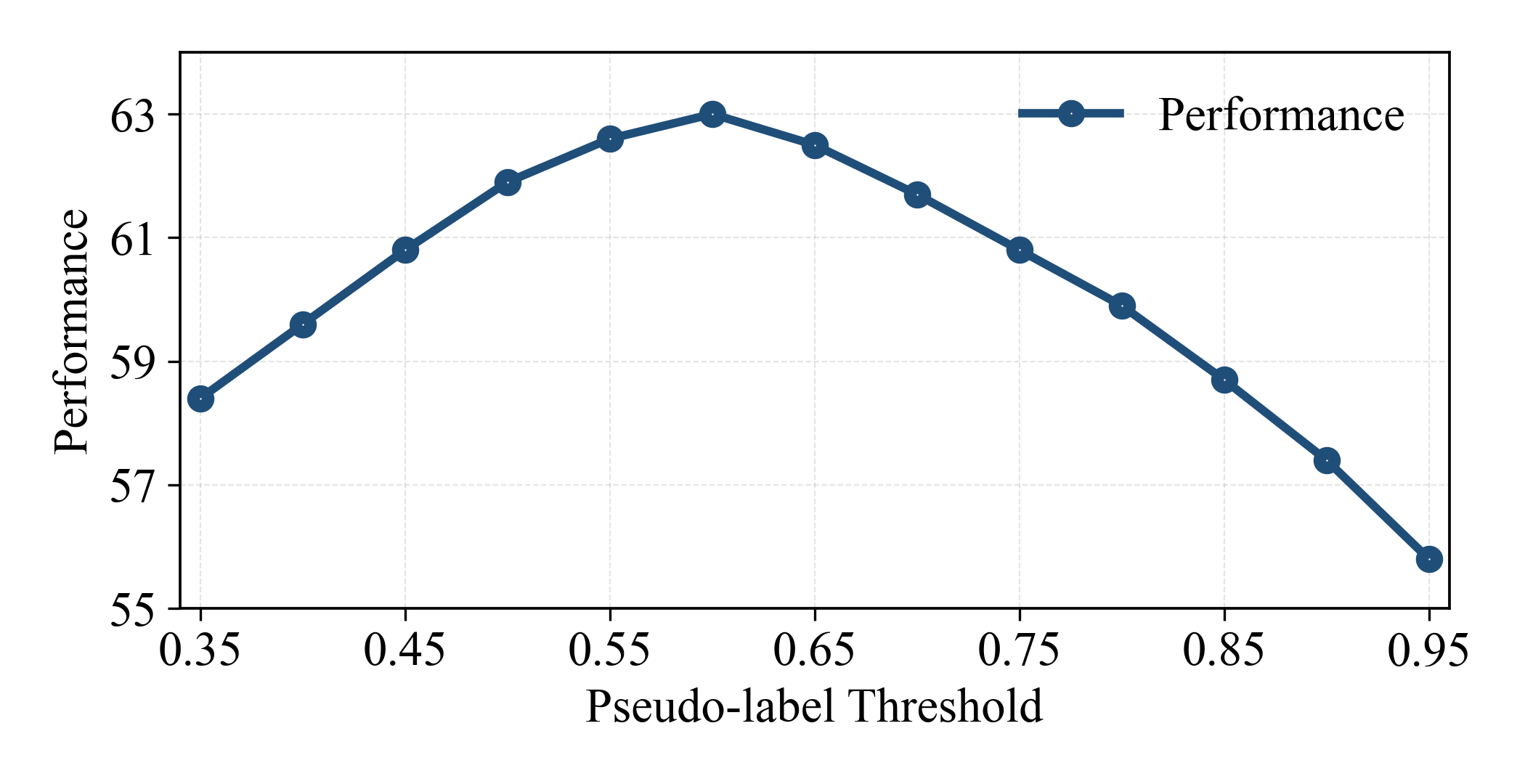}
    \caption{Effect of the pseudo-label confidence threshold. The performance first increases and then decreases as the threshold becomes larger. A small threshold introduces more low-quality pseudo labels, while a large threshold retains too few pseudo boxes for effective training. The best performance is achieved at the threshold of 0.6.}
    \label{fig:pseudo_threshold}
    \vspace{-0.3cm}
\end{figure}

We study the influence of the pseudo-label confidence threshold on detection performance. As illustrated in Fig.~\ref{fig:pseudo_threshold},  the performance reaches its peak when the threshold is set to 0.6 and drops on both sides. A small threshold retains more predicted boxes, but also introduces more low-quality pseudo labels, which may contaminate the training set and harm adaptation. On the other hand, an overly large threshold keeps only a limited number of highly confident boxes, reducing the amount of effective supervision brought by pseudo labeling. These results indicate that a proper threshold is critical for balancing pseudo-label quality and quantity, and 0.6 achieves the best trade-off in our setting.

\subsection{Qualitative Visualization Analysis}

\begin{figure}[t]
    \centering
    \includegraphics[width=\linewidth]{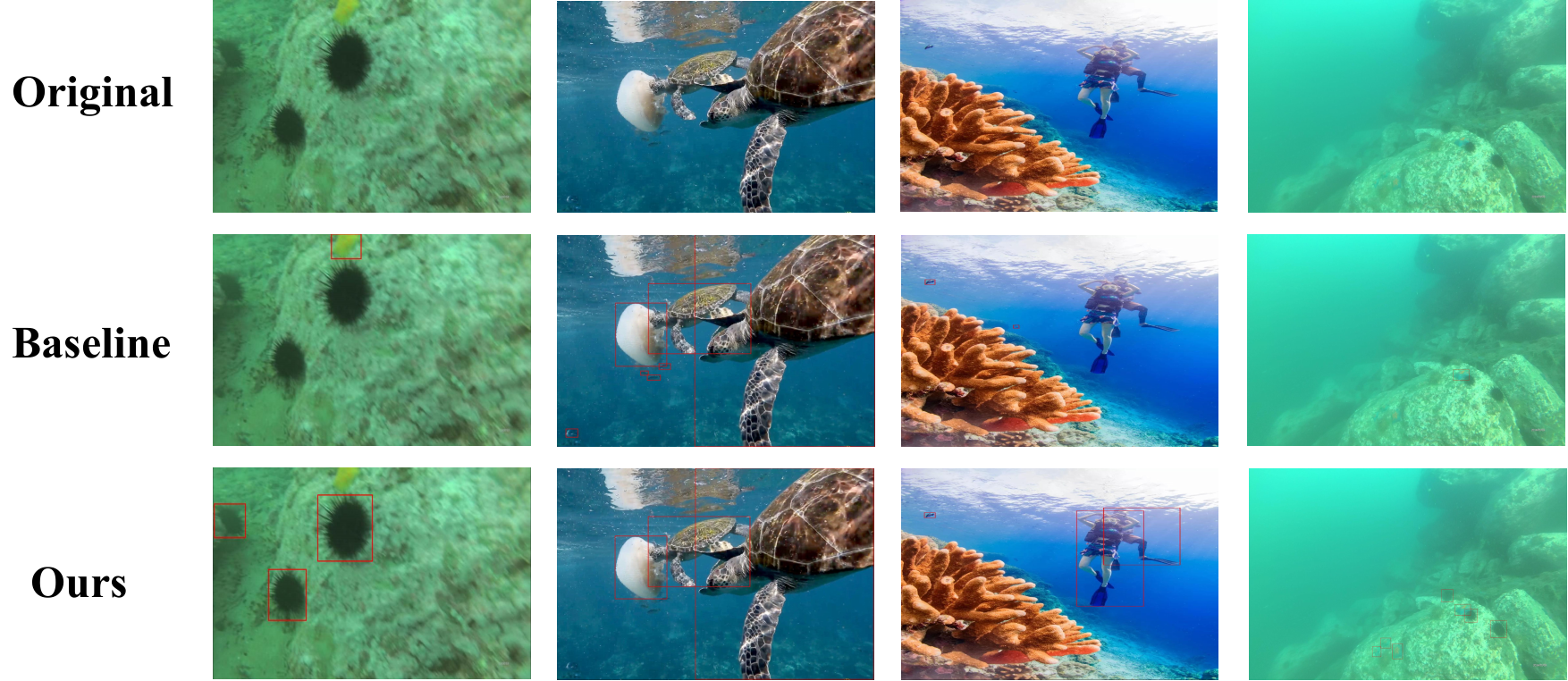}
    \caption{Qualitative comparison of detection results in the underwater target domain. The first row shows the original images, the second row presents the predictions of the baseline, and the third row shows the results of GiPL. Compared with the baseline, GiPL produces more accurate localization, fewer missed detections, and less background interference, demonstrating stronger robustness under challenging cross-domain scenarios.}
    \label{fig:vis_compare}
       \vspace{-0.3cm}
\end{figure}

We further provide qualitative comparisons in Fig.~\ref{fig:vis_compare}. The first row shows the original input images, the second row presents the predictions of the baseline, and the third row shows the results of our GiPL. Compared with the baseline, GiPL produces more accurate and complete detections under challenging underwater scenes. Specifically, the baseline tends to suffer from missed detections, inaccurate localization, and scattered predictions on background regions. In contrast, our method is able to focus more effectively on the true object regions, leading to tighter bounding boxes and fewer distracting responses. This advantage is especially evident in complex cases with blurry appearance, cluttered background, or severe domain shift, where GiPL achieves more reliable foreground localization and better object awareness. These visual results further verify the effectiveness of GiPL in improving target-domain detection quality.

\section{Conclusion}
This work addresses the core bottlenecks of sparse annotation utilization and data scarcity induced overfitting in cross-domain few-shot object detection. We propose GiPL, a two-branch framework that combines iterative pseudo-label self-training and text-guided generative data augmentation. The pseudo-label module fully exploits support set information, while the generative module provides domain-consistent multi-object samples to mitigate overfitting. Experiments on three cross-domain datasets show that our method significantly improves performance under 1/5/10-shot settings and achieves state-of-the-art results in CD-FSOD. In the future, we will extend GiPL to more complex cross-modal and cross-task detection scenarios, and explore lightweight generative augmentation for efficient adaptation.

\section*{Acknowledgments}
This work is supported by the National Natural Science Foundation of China under grants 62206102; the National Key Research and Development Program of China under grant 2024YFC3307900; the National Natural Science Foundation of China under grants 62436003, 62376103 and 62302184; Major Science and Technology Project of Hubei Province under grant 2025BAB011 and 2024BAA008; Hubei Science and Technology Talent Service Project under grant 2024DJC078; and Ant Group through CCF-Ant Research Fund. The computation is completed in the HPC Platform of Huazhong University of Science and Technology.

%% file: main.bbl
\begin{thebibliography}{33}
\providecommand{\natexlab}[1]{#1}
\providecommand{\url}[1]{\texttt{#1}}
\expandafter\ifx\csname urlstyle\endcsname\relax
  \providecommand{\doi}[1]{doi: #1}\else
  \providecommand{\doi}{doi: \begingroup \urlstyle{rm}\Url}\fi

\bibitem[Carion et~al.(2020)Carion, Massa, Synnaeve, Usunier, Kirillov, and Zagoruyko]{carion2020end}
Nicolas Carion, Francisco Massa, Gabriel Synnaeve, Nicolas Usunier, Alexander Kirillov, and Sergey Zagoruyko.
\newblock End-to-end object detection with transformers.
\newblock In \emph{European conference on computer vision}, pages 213--229. Springer, 2020.

\bibitem[Fan et~al.(2020)Fan, Zhuo, Tang, and Tai]{fan2020few}
Qi Fan, Wei Zhuo, Chi-Keung Tang, and Yu-Wing Tai.
\newblock Few-shot object detection with attention-rpn and multi-relation detector.
\newblock In \emph{Proceedings of the IEEE/CVF conference on computer vision and pattern recognition}, pages 4013--4022, 2020.

\bibitem[Fu et~al.(2023{\natexlab{a}})Fu, Liu, Fan, Chen, Fu, Yuan, Zhu, and Luo]{fu2023rethinking}
Chenping Fu, Risheng Liu, Xin Fan, Puyang Chen, Hao Fu, Wanqi Yuan, Ming Zhu, and Zhongxuan Luo.
\newblock Rethinking general underwater object detection: Datasets, challenges, and solutions.
\newblock \emph{Neurocomputing}, 517:\penalty0 243--256, 2023{\natexlab{a}}.

\bibitem[Fu et~al.(2023{\natexlab{b}})Fu, Xie, Fu, and Jiang]{fu2023styleadv}
Yuqian Fu, Yu Xie, Yanwei Fu, and Yu-Gang Jiang.
\newblock Styleadv: Meta style adversarial training for cross-domain few-shot learning.
\newblock In \emph{Proceedings of the IEEE/CVF Conference on Computer Vision and Pattern Recognition}, pages 24575--24584, 2023{\natexlab{b}}.

\bibitem[Fu et~al.(2024)Fu, Wang, Pan, Huai, Qiu, Shangguan, Liu, Fu, Van~Gool, and Jiang]{fu2024cdvito}
Yuqian Fu, Yu Wang, Yixuan Pan, Lian Huai, Xingyu Qiu, Zeyu Shangguan, Tong Liu, Yanwei Fu, Luc Van~Gool, and Xingqun Jiang.
\newblock Cross-domain few-shot object detection via enhanced open-set object detector.
\newblock In \emph{European Conference on Computer Vision}, pages 247--264. Springer, 2024.

\bibitem[Gao et~al.(2022)Gao, Yang, Huang, Xie, Li, and Zheng]{gao2022acrofod}
Yipeng Gao, Lingxiao Yang, Yunmu Huang, Song Xie, Shiyong Li, and Wei-Shi Zheng.
\newblock Acrofod: An adaptive method for cross-domain few-shot object detection.
\newblock In \emph{European Conference on Computer Vision}, pages 673--690. Springer, 2022.

\bibitem[Gao et~al.(2023)Gao, Lin, Yan, Wang, and Zheng]{gao2023asyfod}
Yipeng Gao, Kun-Yu Lin, Junkai Yan, Yaowei Wang, and Wei-Shi Zheng.
\newblock Asyfod: An asymmetric adaptation paradigm for few-shot domain adaptive object detection.
\newblock In \emph{Proceedings of the IEEE/CVF Conference on Computer Vision and Pattern Recognition}, pages 3261--3271, 2023.

\bibitem[Gu et~al.(2021)Gu, Lin, Kuo, and Cui]{gu2021open}
Xiuye Gu, Tsung-Yi Lin, Weicheng Kuo, and Yin Cui.
\newblock Open-vocabulary object detection via vision and language knowledge distillation.
\newblock \emph{arXiv preprint arXiv:2104.13921}, 2021.

\bibitem[Hsieh et~al.(2017)Hsieh, Lin, and Hsu]{Carpk}
Meng{-}Ru Hsieh, Yen{-}Liang Lin, and Winston~H. Hsu.
\newblock Drone-based object counting by spatially regularized regional proposal network.
\newblock In \emph{{IEEE} International Conference on Computer Vision, {ICCV} 2017, Venice, Italy, October 22-29, 2017}, pages 4165--4173. {IEEE} Computer Society, 2017.

\bibitem[Kang et~al.(2019)Kang, Liu, Wang, Yu, Feng, and Darrell]{kang2019few}
Bingyi Kang, Zhuang Liu, Xin Wang, Fisher Yu, Jiashi Feng, and Trevor Darrell.
\newblock Few-shot object detection via feature reweighting.
\newblock In \emph{Proceedings of the IEEE/CVF international conference on computer vision}, pages 8420--8429, 2019.

\bibitem[Lee et~al.(2022)Lee, Yang, Chakraborty, Cai, Swaminathan, Ravichandran, and Dabeer]{lee2022mofsod}
Kibok Lee, Hao Yang, Satyaki Chakraborty, Zhaowei Cai, Gurumurthy Swaminathan, Avinash Ravichandran, and Onkar Dabeer.
\newblock Rethinking few-shot object detection on a multi-domain benchmark.
\newblock In \emph{European Conference on Computer Vision}, pages 366--382. Springer, 2022.

\bibitem[Li et~al.(2022)Li, Zhang, Zhang, Yang, Li, Zhong, Wang, Yuan, Zhang, Hwang, et~al.]{li2022grounded}
Liunian~Harold Li, Pengchuan Zhang, Haotian Zhang, Jianwei Yang, Chunyuan Li, Yiwu Zhong, Lijuan Wang, Lu Yuan, Lei Zhang, Jenq-Neng Hwang, et~al.
\newblock Grounded language-image pre-training.
\newblock In \emph{Proceedings of the IEEE/CVF conference on computer vision and pattern recognition}, pages 10965--10975, 2022.

\bibitem[Li et~al.(2025)Li, Qiu, Fu, Chen, Qian, Zheng, Paudel, Fu, Huang, Van~Gool, et~al.]{li2025domain}
Yu Li, Xingyu Qiu, Yuqian Fu, Jie Chen, Tianwen Qian, Xu Zheng, Danda~Pani Paudel, Yanwei Fu, Xuanjing Huang, Luc Van~Gool, et~al.
\newblock Domain-rag: Retrieval-guided compositional image generation for cross-domain few-shot object detection.
\newblock \emph{arXiv preprint arXiv:2506.05872}, 2025.

\bibitem[Liu et~al.(2024)Liu, Zeng, Ren, Li, Zhang, Yang, Jiang, Li, Yang, Su, et~al.]{liu2024grounding}
Shilong Liu, Zhaoyang Zeng, Tianhe Ren, Feng Li, Hao Zhang, Jie Yang, Qing Jiang, Chunyuan Li, Jianwei Yang, Hang Su, et~al.
\newblock Grounding dino: Marrying dino with grounded pre-training for open-set object detection.
\newblock In \emph{European conference on computer vision}, pages 38--55. Springer, 2024.

\bibitem[Pan et~al.(2025{\natexlab{a}})Pan, Liu, Fu, Ma, Li, Paudel, Gool, and Huang]{LAE-DINO}
Jiancheng Pan, Yanxing Liu, Yuqian Fu, Muyuan Ma, Jiahao Li, Danda~Pani Paudel, Luc~Van Gool, and Xiaomeng Huang.
\newblock Locate anything on earth: Advancing open-vocabulary object detection for remote sensing community.
\newblock In \emph{Thirty-Ninth {AAAI} Conference on Artificial Intelligence, Thirty-Seventh Conference on Innovative Applications of Artificial Intelligence, Fifteenth Symposium on Educational Advances in Artificial Intelligence, {AAAI} 2025, Philadelphia, PA, USA, February 25 - March 4, 2025}, pages 6281--6289. {AAAI} Press, 2025{\natexlab{a}}.

\bibitem[Pan et~al.(2025{\natexlab{b}})Pan, Liu, He, Peng, Li, Sun, and Huang]{pan2025enhance}
Jiancheng Pan, Yanxing Liu, Xiao He, Long Peng, Jiahao Li, Yuze Sun, and Xiaomeng Huang.
\newblock Enhance then search: An augmentation-search strategy with foundation models for cross-domain few-shot object detection.
\newblock In \emph{CVPRW}, 2025{\natexlab{b}}.

\bibitem[Qiu et~al.(2026)Qiu, Fu, Jiawei, Ren, Pan, Fu, Timofte, et~al.]{qiu2026ntire}
Xingyu Qiu, Yuqian Fu, Geng Jiawei, Bin Ren, Jiancheng Pan, Yanwei Fu, Radu Timofte, et~al.
\newblock Ntire 2026 challenge on cross-domain few-shot object detection: methods and results.
\newblock In \emph{CVPRW}, 2026.

\bibitem[Redmon and Farhadi(2018)]{redmon2018yolov3}
Joseph Redmon and Ali Farhadi.
\newblock Yolov3: An incremental improvement.
\newblock \emph{arXiv preprint arXiv:1804.02767}, 2018.

\bibitem[Redmon et~al.(2016)Redmon, Divvala, Girshick, and Farhadi]{redmon2016you}
Joseph Redmon, Santosh Divvala, Ross Girshick, and Ali Farhadi.
\newblock You only look once: Unified, real-time object detection.
\newblock In \emph{Proceedings of the IEEE conference on computer vision and pattern recognition}, pages 779--788, 2016.

\bibitem[Ren et~al.(2015)Ren, He, Girshick, and Sun]{ren2015faster}
Shaoqing Ren, Kaiming He, Ross Girshick, and Jian Sun.
\newblock Faster r-cnn: Towards real-time object detection with region proposal networks.
\newblock \emph{Advances in neural information processing systems}, 28, 2015.

\bibitem[Sun et~al.(2021)Sun, Li, Cai, Yuan, and Zhang]{sun2021fsce}
Bo Sun, Banghuai Li, Shengcai Cai, Ye Yuan, and Chi Zhang.
\newblock Fsce: Few-shot object detection via contrastive proposal encoding.
\newblock In \emph{Proceedings of the IEEE/CVF conference on computer vision and pattern recognition}, pages 7352--7362, 2021.

\bibitem[Wang et~al.(2020)Wang, Huang, Darrell, Gonzalez, and Yu]{wang2020frustratingly}
Xin Wang, Thomas~E Huang, Trevor Darrell, Joseph~E Gonzalez, and Fisher Yu.
\newblock Frustratingly simple few-shot object detection.
\newblock \emph{arXiv preprint arXiv:2003.06957}, 2020.

\bibitem[Wang et~al.(2023)Wang, Li, and Wu]{CarDD}
Xinkuang Wang, Wenjing Li, and Zhongcheng Wu.
\newblock Cardd: {A} new dataset for vision-based car damage detection.
\newblock \emph{{IEEE} Trans. Intell. Transp. Syst.}, 24\penalty0 (7):\penalty0 7202--7214, 2023.

\bibitem[Wu et~al.(2025)Wu, Li, Zhou, Lin, Gao, Yan, Yin, Bai, Xu, Chen, Chen, Tang, Zhang, Wang, Yang, Yu, Cheng, Liu, Li, Zhang, Meng, Wei, Ni, Chen, Cao, Peng, Qu, Wu, Wang, Yu, Wen, Feng, Xu, Wang, Zhang, Zhu, Wu, Cai, and Liu]{wu2025qwen}
Chenfei Wu, Jiahao Li, Jingren Zhou, Junyang Lin, Kaiyuan Gao, Kun Yan, Shengming Yin, Shuai Bai, Xiao Xu, Yilei Chen, Yuxiang Chen, Zecheng Tang, Zekai Zhang, Zhengyi Wang, An Yang, Bowen Yu, Chen Cheng, Dayiheng Liu, Deqing Li, Hang Zhang, Hao Meng, Hu Wei, Jingyuan Ni, Kai Chen, Kuan Cao, Liang Peng, Lin Qu, Minggang Wu, Peng Wang, Shuting Yu, Tingkun Wen, Wensen Feng, Xiaoxiao Xu, Yi Wang, Yichang Zhang, Yongqiang Zhu, Yujia Wu, Yuxuan Cai, and Zenan Liu.
\newblock Qwen-image technical report.
\newblock \emph{arXiv preprint arXiv:2508.02324}, 2025.

\bibitem[Xiong(2023)]{xiong2023cd}
Wuti Xiong.
\newblock Cd-fsod: A benchmark for cross-domain few-shot object detection.
\newblock In \emph{ICASSP 2023-2023 IEEE International Conference on Acoustics, Speech and Signal Processing (ICASSP)}, pages 1--5. IEEE, 2023.

\bibitem[Yan et~al.(2019)Yan, Chen, Xu, Wang, Liang, and Lin]{yan2019meta}
Xiaopeng Yan, Ziliang Chen, Anni Xu, Xiaoxi Wang, Xiaodan Liang, and Liang Lin.
\newblock Meta r-cnn: Towards general solver for instance-level low-shot learning.
\newblock In \emph{Proceedings of the IEEE/CVF international conference on computer vision}, pages 9577--9586, 2019.

\bibitem[Zareian et~al.(2021)Zareian, Rosa, Hu, and Chang]{zareian2021open}
Alireza Zareian, Kevin~Dela Rosa, Derek~Hao Hu, and Shih-Fu Chang.
\newblock Open-vocabulary object detection using captions.
\newblock In \emph{Proceedings of the IEEE/CVF conference on computer vision and pattern recognition}, pages 14393--14402, 2021.

\bibitem[Zhang et~al.(2022)Zhang, Li, Liu, Zhang, Su, Zhu, Ni, and Shum]{zhang2022dino}
Hao Zhang, Feng Li, Shilong Liu, Lei Zhang, Hang Su, Jun Zhu, Lionel~M Ni, and Heung-Yeung Shum.
\newblock Dino: Detr with improved denoising anchor boxes for end-to-end object detection.
\newblock \emph{arXiv preprint arXiv:2203.03605}, 2022.

\bibitem[Zhao et~al.(2026)Zhao, Zou, Li, and Li]{zhao2026interpretable}
Yaze Zhao, Yixiong Zou, Yuhua Li, and Ruixuan Li.
\newblock Interpretable cross-domain few-shot learning with rectified target-domain local alignment.
\newblock \emph{arXiv preprint arXiv:2603.17655}, 2026.

\bibitem[Zhu et~al.(2020)Zhu, Su, Lu, Li, Wang, and Dai]{zhu2020deformable}
Xizhou Zhu, Weijie Su, Lewei Lu, Bin Li, Xiaogang Wang, and Jifeng Dai.
\newblock Deformable detr: Deformable transformers for end-to-end object detection.
\newblock \emph{arXiv preprint arXiv:2010.04159}, 2020.

\bibitem[Zou et~al.(2024{\natexlab{a}})Zou, Liu, Hu, Li, and Li]{DBLP:conf/cvpr/ZouLH0024}
Yixiong Zou, Yicong Liu, Yiman Hu, Yuhua Li, and Ruixuan Li.
\newblock Flatten long-range loss landscapes for cross-domain few-shot learning.
\newblock In \emph{{CVPR} 2024}, pages 23575--23584. {IEEE}, 2024{\natexlab{a}}.

\bibitem[Zou et~al.(2024{\natexlab{b}})Zou, Ma, Li, and Li]{zou2024attention}
Yixiong Zou, Ran Ma, Yuhua Li, and Ruixuan Li.
\newblock Attention temperature matters in vit-based cross-domain few-shot learning.
\newblock In \emph{NeurIPS 2024}, 2024{\natexlab{b}}.

\bibitem[Zou et~al.(2024{\natexlab{c}})Zou, Yi, Li, and Li]{zou2024a}
Yixiong Zou, Shuai Yi, Yuhua Li, and Ruixuan Li.
\newblock A closer look at the {CLS} token for cross-domain few-shot learning.
\newblock In \emph{NeurIPS 2024}, 2024{\natexlab{c}}.

\end{thebibliography}
